\documentclass[10pt,twocolumn,letterpaper]{article}

\usepackage{cvpr}              

\usepackage{graphicx}
\usepackage{amsmath}
\usepackage{amssymb}
\usepackage{booktabs}

\usepackage{algorithm}
\usepackage{algorithmic}

\usepackage{caption}
\usepackage{subcaption}
\usepackage{array}
\usepackage{colortbl}
\usepackage{booktabs}
\usepackage{multirow}
\usepackage{multicol}
\usepackage{makecell}
\usepackage{xcolor}
\usepackage{xspace}
\usepackage{float}
\usepackage{color}
\usepackage{bbding}
\usepackage{pifont}
\usepackage{enumitem} 

\usepackage{url}
\usepackage{listings} 

\usepackage[pagebackref,breaklinks,colorlinks]{hyperref}

\definecolor{mygreen}{RGB}{83,161,81}
\definecolor{myred}{RGB}{178,34,34}
\definecolor{lightorange}{RGB}{249,195,129}
\definecolor{borderblue}{RGB}{71,117,194}
\definecolor{borderyellow}{RGB}{253,190,38}
\newcommand{\posacc}[1]{{\bf \fontsize{8.0}{42}\selectfont \color{mygreen}~(#1)}}
\newcommand{\negacc}[1]{{\bf \fontsize{8.0}{42}\selectfont \color{myred}~(#1)}}
\newcommand{\myparagraph}[1]{{\vspace{.5em} \noindent \bf #1}}

\usepackage[capitalize]{cleveref}
\crefname{section}{Sec.}{Secs.}
\Crefname{section}{Section}{Sections}
\Crefname{table}{Table}{Tables}
\crefname{table}{Tab.}{Tabs.}

\def\cvprPaperID{xxxx} 
\def\confName{CVPR}
\def\confYear{2022}

\begin{document}

\title{Language as Queries for Referring Video Object Segmentation}

\author
{
Jiannan Wu$^{1}$, 
~~~
Yi Jiang$^{2}$
~~~
Peize Sun$^{1}$, 
~~~
Zehuan Yuan$^{2}$, 
~~~
Ping Luo$^{1}$
\\[0.2cm]
${^1}$The University of Hong Kong ~~~
${^2}$ByteDance
}


\maketitle

\begin{abstract}
   Referring video object segmentation (R-VOS) is an emerging cross-modal task that aims to segment the target object referred by a language expression in all video frames. 
   In this work, we propose a simple and unified framework built upon Transformer, termed ReferFormer. It views the language as queries and directly attends to the most relevant regions in the video frames. Concretely, we introduce a small set of object queries conditioned on the language as the input to the Transformer. In this manner, all the queries are obligated to find the referred objects only. They are eventually transformed into dynamic kernels which capture the crucial object-level information, and play the role of convolution filters to generate the segmentation masks from feature maps. The object tracking is achieved naturally by linking the corresponding queries across frames. This mechanism greatly simplifies the pipeline and the end-to-end framework is significantly different from the previous methods. 
   Extensive experiments on Ref-Youtube-VOS, Ref-DAVIS17, A2D-Sentences and JHMDB-Sentences show the effectiveness of ReferFormer. On Ref-Youtube-VOS, ReferFormer achieves 55.6 $\mathcal{J}\&\mathcal{F}$ with a ResNet-50 backbone without bells and whistles, which exceeds the previous state-of-the-art performance by 8.4 points. In addition, with the strong Swin-Large backbone, ReferFormer achieves the best $\mathcal{J}\&\mathcal{F}$ of 64.2 among all existing methods. Moreover, we show the impressive results of 55.0 mAP and 43.7 mAP on A2D-Sentences and JHMDB-Sentences respectively, which significantly outperforms the previous methods by a large margin.
   Code is publicly available at \href{https://github.com/wjn922/ReferFormer}{https://github.com/wjn922/ReferFormer}.
\end{abstract}


\section{Introduction}

\begin{figure}[t]
\begin{center}
   \includegraphics[width=1.0\linewidth]{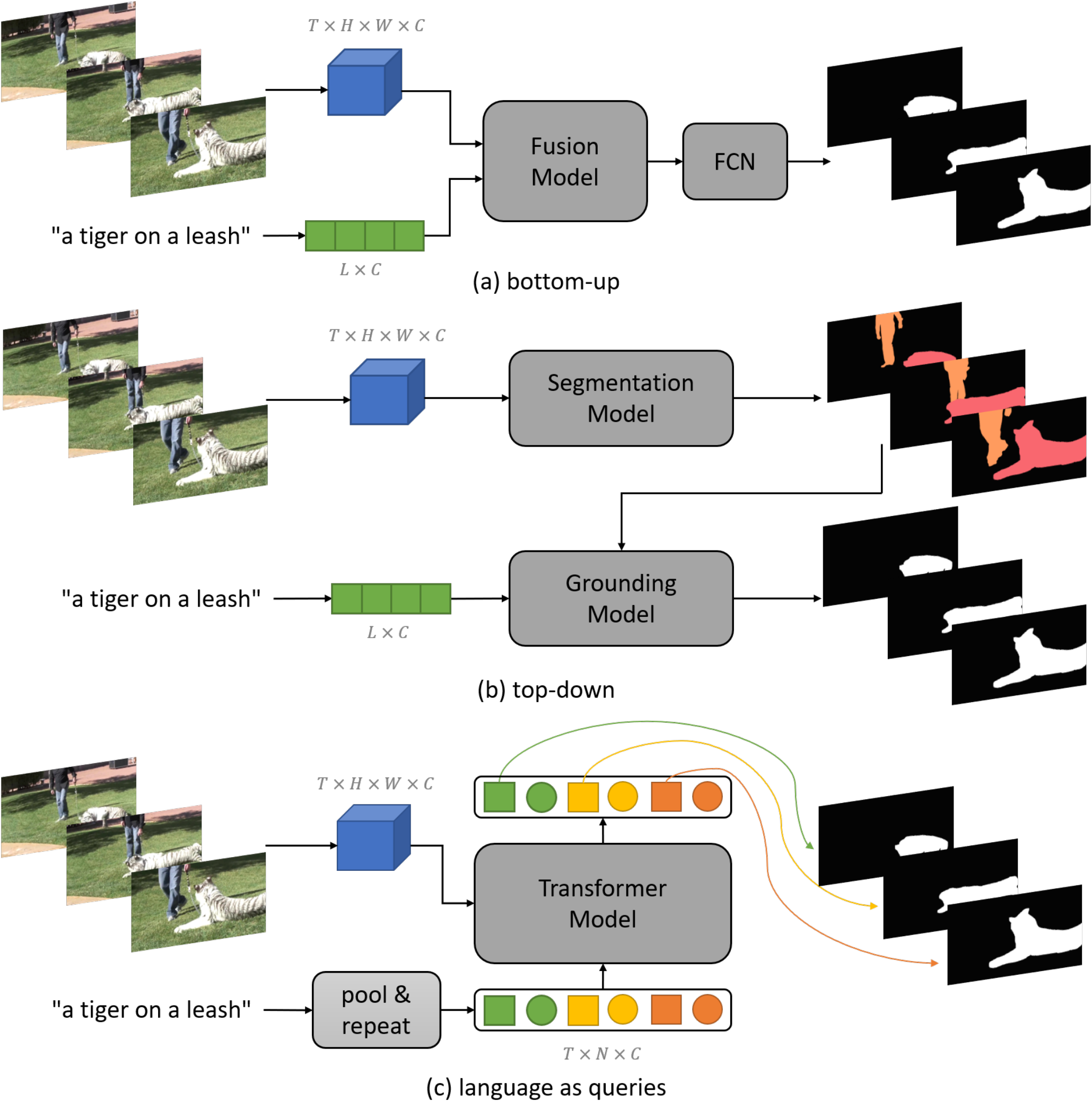}
\end{center}
\vspace{-4mm}
   \caption{Comparison of current referring video object segmentation (R-VOS) pipelines. \textbf{(a)} Bottom-up. \textbf{(b)} Top-down. \textbf{(c)} Ours.}
\label{fig:compare}
\vspace{-5mm}
\end{figure}

Referring video object segmentation (R-VOS) aims to segment the target object in a video given a natural language description. This emerging topic has raised great attention in the research community and is expected to benefit many applications in a friendly and interactive way, \eg, video editing and video surveillance. R-VOS is more challenging than the traditional semi-supervised video object segmentation ~\cite{perazzi2016davis, xu2018youtube}, because it does not only lack the ground-truth mask annotation in the first frame, but also require the comprehensive understanding of the cross-modal sources, \ie, vision and language. Therefore, the model should have a strong ability to infer which object is referred and to perform accurate segmentation.

To accomplish this task, the existing methods can be mainly categorized into two groups: (1) \textit{Bottom-up} methods. These methods incorporate the vision and language features in a early-fusion manner, and then adopt a FCN ~\cite{long2015fcn} as decoder to generate object masks, as shown in Figure \ref{fig:compare}(a). (2) \textit{Top-down} methods. These methods tackle the problem in a top-down perspective and follow a two-stage pipeline. As illustrated in Figure \ref{fig:compare}(b), they first employ an instance segmentation model to find all the objects in each frame, and then associate them in the entire video to form the tracklet candidates. Afterwards, they use the expression as the grounding criterion to select the best-matched one.

Although these two streams of methods have demonstrated their effectiveness with promising results, they still have some intrinsic limitations. First, for the \textit{bottom-up} methods, they fail to capture the crucial instance-level information and do not consider the object association across multiple frames. Therefore, this type of methods can not provide explicit knowledge for cross-modal reasoning and would encounter the discrepancy of predicted object due to scene changes. Second, although \textit{top-down} methods have greatly boost the performance over the \textit{bottom-up} methods, they suffer from heavy workload because of the complex, multi-stage pipeline. For example, the recent method proposed by Liang \etal ~\cite{liang2021topdown} comprises of three parts: HTC~\cite{chen2019htc}, CFBI~\cite{yang2020cfbi} and a tracklet-language grounding model. All these networks need to be pretrained on the ImageNet ~\cite{krizhevsky2012imagenet}, COCO ~\cite{lin2014coco} or RefCOCO ~\cite{yu2016refcoco} and further finetuned on R-VOS datasets, respectively. Furthermore, the separate optimization on several sub-problems would lead to sub-optimal solution.

These limitations of current methods motivate us to design a simple and unified framework that solves the R-VOS task elegantly. The recent success of Transformer ~\cite{vaswani2017transformer} in object detection ~\cite{carion2020detr, zhu2020deformable} and video instance segmentation ~\cite{wang2021vistr, hwang2021ifc, wu2021seqformer} demonstrates a promising solution. However, it is non-trivial to apply such models to the R-VOS task. These models ~\cite{carion2020detr, zhu2020deformable} use a fixed number (\eg, 100) of learnable queries to detect all the objects in an image. Under this circumstance, it would be confused for the model to distinguish which object is referred due to the randomness of the expression. 
Here raises a natural question: "\textit{Is it possible for a unified model to know where to look using queries?"}

This work answers the question by proposing the notion of \textit{language as queries}, as shown in Figure \ref{fig:compare}(c). We put the linguistic restriction on all object queries and use these \textit{conditional queries} as input for the model. In this manner, the expression will make the queries focus on the referred object only, and thus greatly reducing the query number (\eg, 5 in our experiments). The next challenge lies in how to decode the object mask from query representations. As the queries contain rich instance characteristics, we view them as instance-aware dynamic kernels to filter out the segmentation masks from feature maps. Moreover, to make the feature maps more discriminative, we design a novel cross-modal feature pyramid network (CM-FPN) where the visual and linguistic features interact in multiple levels for fine-grained cross-modal fusion. 

The unified framework can not only produce the segmentation masks for referred objects, but also the classification results and detection boxes. Moreover, the \textit{conditional queries} are linked via instance matching strategy across frames so that the object tracking is achieved naturally without post-process. As shown in Figure \ref{fig:visualize}, our unified framework is able to detect, segment and track the referred object simultaneously. We hope this framework could serve as a strong baseline for R-VOS task.

The main contributions of this work are as follows.

\begin{itemize}[leftmargin=*]
\vspace{-2mm}
\item We propose a simple and unified framework for referring video object segmentation, termed ReferFormer. Given a video clip and the corresponding language expression, our framework directly detects, segments and tracks the referred object in all frames in an end-to-end manner.
\vspace{-2mm}
\item We present the notion of \textit{language as queries}. We introduce a small set of object queries which conditioned on the text expression to attend the referred object only. These \textit{conditional queries} are shared across different frames in the initial state and they are transformed into dynamic kernels to filter out the segmentation masks from feature maps. This mechanism provides a new perspective for the R-VOS task.
\vspace{-2mm}
\item We design the cross-modal feature pyramid network (CM-FPN) for multi-scale vision-language fusion, which improves the discriminativeness of mask features for accurate segmentation.
\vspace{-2mm}
\item Extensive experiments on Ref-Youtube-VOS, Ref-DAVIS17, A2D-Sentences and JHMDB-Sentences show that ReferFormer outperforms the previous methods on these four benchmarks by a large margin. \Eg, on Ref-Youtube-VOS, ReferFormer with a ResNet-50 backbone achieves 55.6 $\mathcal{J}\&\mathcal{F}$ without bells and whistles, showing the significant 8.4 points gain over the previous state-of-the-art methods. And using the strong Video-Swin-Base visual backbone, ReferFormer achieves the impressive results of 64.9 $\mathcal{J}\&\mathcal{F}$.
\end{itemize}

\section{Related Work}


\myparagraph{Semi-supervised Video Object Segmentation.} The traditional semi-supervised video object segmentation (Semi-VOS) aims to propagates the ground-truth object masks given in the first frame to the entire video. Most recent works ~\cite{voigtlaender2019feelvos, oh2019stm, cheng2021stcn, yang2020cfbi} lie in the group of matching-based methods, which perform feature matching to track the target objects. STM ~\cite{oh2019stm} leverages a memory to store the past object features and utilize the attention matching mechanism on the memory to guide the prediction of current frame. CFBI ~\cite{yang2020cfbi} not only considers the embedding learning of foreground objects but also the background, resulting in a more robust framework.

\begin{figure*}[ht]
\begin{center}
\includegraphics[width=0.85\textwidth]{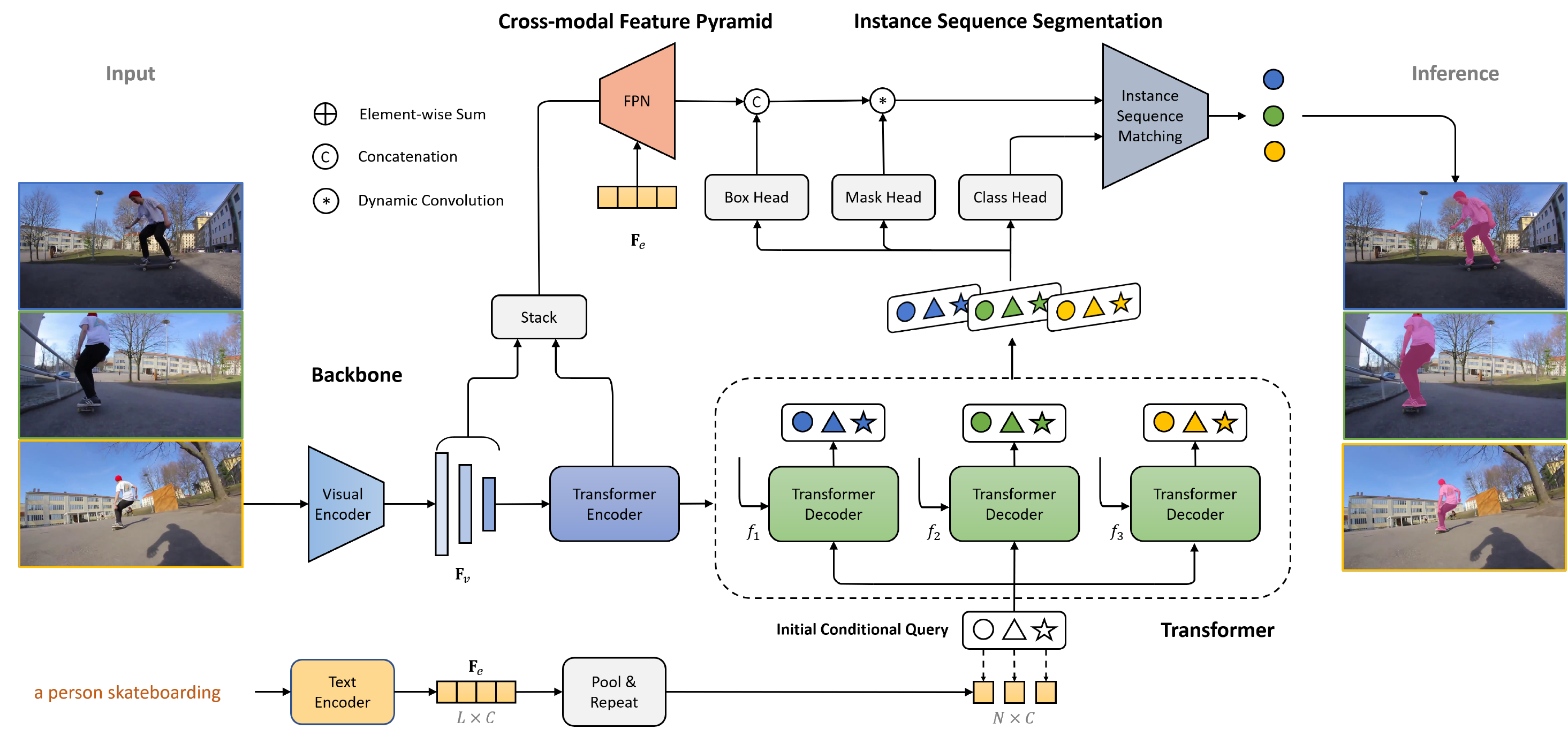}
\end{center}
\vspace{-4mm}
\caption{The overall pipeline of ReferFormer. It mainly consists of four parts: Backbone, Transformer, Cross-modal Feature Pyramid and the Segmentation part. The model takes a video clip with the corresponding language expression as input and output the segmentation mask of the referred object in each frame. For the Transformer decoder input, the object queries are conditioned on the language expression to find the referred object. The same colors represent the queries in the same frame and the same shapes represent the queries refer to the same instance. The order of queries inner frame keep consistent for different frames. Best viewed in color.}
\label{fig:overall}
\vspace{-4mm}
\end{figure*}

\myparagraph{Referring Video Object Segmentation.} 
Referring video object segmentation (R-VOS) provides the language description instead of mask annotation as the object reference, thus it would be a more challenging task. The current methods for R-VOS mainly follow the two pipelines: (1) \textit{Bottom-up} methods. An intuitive thinking is directly applying the image-level methods ~\cite{luo2020mcn, huang2020cmpc, zhang2017dbnet, ding2021vlt} on the video frames independently, \eg, RefVOS ~\cite{bellver2020refvos}. The obvious drawback of such methods is that they fail to utilize the valuable temporal information across frames, resulting in inconsistent object prediction due to the scene or appearance variations. To address this issue, URVOS ~\cite{seo2020urvos} casts the task as a joint problem of referring object segmentation in an image and mask propagation in a video. They propose a unified referring VOS framework that employs a memory attention module to leverage the information of mask predictions in previous frames. (2) \textit{Top-down} methods. The typical top-down method ~\cite{liang2021topdown} first constructs an exhaustive set of object tracklets by propagating the object masks detected from several key frames to the whole video. Then, a language grounding model is built to select the best object tracklet from the candidate set. Although the method has made breakthrough performance improvement over the previous methods, the complex, multi-stage pipeline is computational-expensive and impractical. 
\vspace{-1mm}

In contrast to these two pipelines, we propose a \textit{query-based} method that achieves the strongest performance with a simple and unified framework. The very recent work MTTR ~\cite{botach2021mttr} also relies on the query-based mechanism. Nevertheless, they need the exhaustive segmentation annotations of all objects and supervised the un-referred instances during training process, which increases the workload of laborious annotation and makes the framework limited in practical applications.
\vspace{-2mm}

\myparagraph{Transformer}
Transformer ~\cite{vaswani2017transformer} was first introduced for sequence-to-sequence translation in natural language processing (NLP) community and has achieved marvelous success in most computer vision tasks~\cite{dosovitskiy2020vit, liu2021swin, kamath2021mdetr} such as object detection\cite{carion2020detr, zhu2020deformable}, tracking\cite{sun2020transtrack,Trackformer,TransT,STARK} and segmentation\cite{SETR,cheng2021maskformer,hwang2021ifc}. DETR ~\cite{carion2020detr} introduces the new \textit{query-based} paradigm\cite{zhu2020deformable, sun2021sparsercnn} for object detection, which employs a set of object queries as candidates and inputs them to the Transformer decoder. Beyond image field, VisTR ~\cite{wang2021vistr} extends the framework for video instance segmentation (VIS) ~\cite{yang2019vis} task and solves the problem in a direct end-to-end parallel sequence decoding manner. 
SeqFormer ~\cite{wu2021seqformer} decouples the content query and box query to aggregates temporal information from each frame and achieves the state-of-the-art performance on VIS task. Inspired by these works, our work also relies on the \textit{query-based} mechanism of Transformer but considers an additional modality, \ie, language, as the object reference. Thus, we propose the notion of \textit{language as queries} and build the simple and unified framework that detects, segments and tracks the referred object simultaneously.

\section{Approach}

Given a video clip $\mathcal{I}= \left \{ I_{t} \right \}_{t=1}^{T}$ with $T$ frames and a referring expression $\mathcal{E} = \left \{ e_{l} \right \}_{l=1}^{L}$ with $L$ words, we aim to produce $T$-frame binary segmentation masks of referred object $\mathcal{S}= \left \{ s_{t} \right \}_{t=1}^{T}, s_{t} \in \mathbb{R}^{H \times W}$ in an end-to-end manner. To this end, we propose a simple and unified framework named ReferFormer, as shown in Figure \ref{fig:overall}.
It mainly consists of four key components: Backbone, Transformer, Cross-modal Feature Pyramid network (CM-FPN) and the Instance Sequence Segmentation process. A small set of object queries conditioned on the language is introduced to find the referred object.
During inference, we directly output the mask predictions by selecting the queries with the highest average score as the final results. 

\subsection{Backbone} \label{sec:backbone}

\myparagraph{Visual Encoder.} We start by adopting a visual backbone to extract the multi-scale feature maps for each frame in the video clip independently, resulting in the visual feature sequence $\mathcal{F}_{v} = \left \{ f_{t} \right \}_{t=1}^{T}$. It is noteworthy that both the 2D spatial encoder (\eg, ResNet ~\cite{he2016resnet}) and 3D spatio-temporal encoder (\eg, Video Swin Transformer ~\cite{liu2021videoswin} could play the role of visual backbone. 

\vspace{-2mm}

\myparagraph{Linguistic Encoder.} Given the language description with $L$ words, we use off-the-shelf linguistic embedding model, RoBERTa ~\cite{liu2019roberta}, to extract the text feature $\mathcal{F}_{e} = \left \{ f_{i} \right \}_{i=1}^{L}$. And we also obtain the sentence-level feature $f_{e}^{s} \in \mathbb{R}^{C}$ by pooling the features of each word. They are both necessary and essential in our model, because the sentence feature guides the learnable queries to find the referred object and text features will have fine-grained interaction with the visual features for reliable cross-modal reasoning.

\subsection{Language as Queries} \label{sec:language_as_queries}

The key design comes from that we use a set of object queries conditioned on the language expression, termed \textit{conditional queries}, as the Transformer decoder input. These queries are obligated to focus on the referred object only and produce the instance-aware dynamic kernels. The final segmentation masks are obtained by performing dynamic convolution between the dynamic kernels and their corresponding feature maps. Here, we adopt the Deformable-DETR ~\cite{zhu2020deformable} as our Transformer model due to its effectiveness and efficiency to capture the global pixel-level relations. 

\myparagraph{Transformer Encoder.} First, a 1 $\times$ 1 convolution is applied on the multi-scale visual features $\mathcal{F}_{v}$ to reduce the channel dimension of all feature maps to $C = 256$. To enrich the information of visual features, we then incorporate projected visual features with the text feature $\mathcal{F}_{e}$ in a multiplication way and form the new multi-scale feature maps $\mathcal{F}_{v}^{'} = \left \{ f_{t}^{'} \right \}_{t=1}^{T}$. Afterwards, the fixed 2D positional encoding is added to feature maps of each frame and the summed features are fed into the Transformer encoder. To utilize the Transformer process the video frames independently, we flatten the spatial dimensions and move the temporal dimension to batch dimension for efficiency. Finally, the output of the Transformer encoder, \ie, \textit{encoded memory}, is then input to the decoder.

\begin{figure}[t]
\begin{center}
   \includegraphics[width=1.0\linewidth]{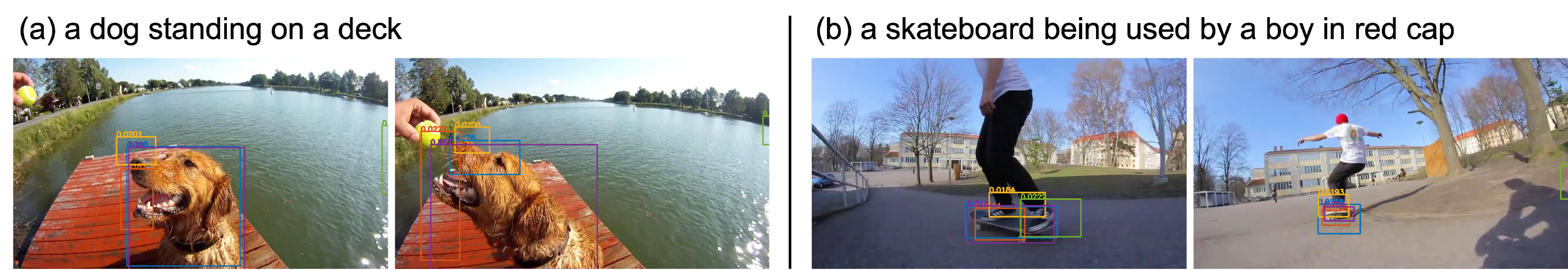}
\end{center}
\vspace{-4mm}
  \caption{We visualize the predicted boxes from all the queries. It can be seen that the these boxes will locate near the referred object only even if there are other objects in the video.}
\label{fig:vis_box}
\vspace{-2mm}
\end{figure}

\myparagraph{Transformer Decoder.} We introduce $N$ object queries to represent the instances for each frame similar to ~\cite{wang2021vistr}, the difference lies in that the query weights are shared across video frames. This mechanism is more flexible to handle the length-variable videos and is more robust for the queries to track the same instances. Meanwhile, we repeat the sentence feature $f_{e}^{s}$ for $N$ times to fit the query number. Both the object queries and repeated sentence features are fed into the decoder as input. In this manner, all the queries will use the language expression as guidance and try to find the referred objects only. These \textit{conditional queries} are duplicated to serve as the decoder input for all the frames and they are turned into instance embeddings by the decoder eventually, resulting in the set of $N_{q} = T \times N$ predictions. It should be noted the queries keep the same order across different frames and we refer to the queries in the same relative position (represented as the same shape in Figure \ref{fig:overall}) as \textit{instance sequence} following ~\cite{wang2021vistr}. Therefore, the temporal coherence of referred object could be achieved easily by linking the corresponding queries.

\myparagraph{Prediction Heads.} Three lightweight heads are built on top of the decoder to further transform the $N_{q}$ instance embeddings. The class head outputs the binary probability which indicates whether the instance is referred by the text sentence and this instance is visible in the current frame. It could also be modified to predict the referred object category by simply changing the output class number. The mask head is implemented by three consecutive linear layers. It produces the parameters of $N_{q}$ dynamic kernels $\Omega = \left \{ \omega_{i} \right \}_{i=1}^{N_{q}}$, which is similar to the conditional convolutional filters in ~\cite{tian2020condist}. These parameters will be reshaped to form the three $1 \times 1$ convolution layers with the channel number as 8. The box head is a 3-layer feed forward network (FFN) with ReLU activation except for the last layer. It will predict the box location of the referred object and thus the position of dynamic kernels could be determined by the center of corresponding boxes. 

\begin{figure*}[t]
\begin{center}
   \includegraphics[width=1.0\linewidth]{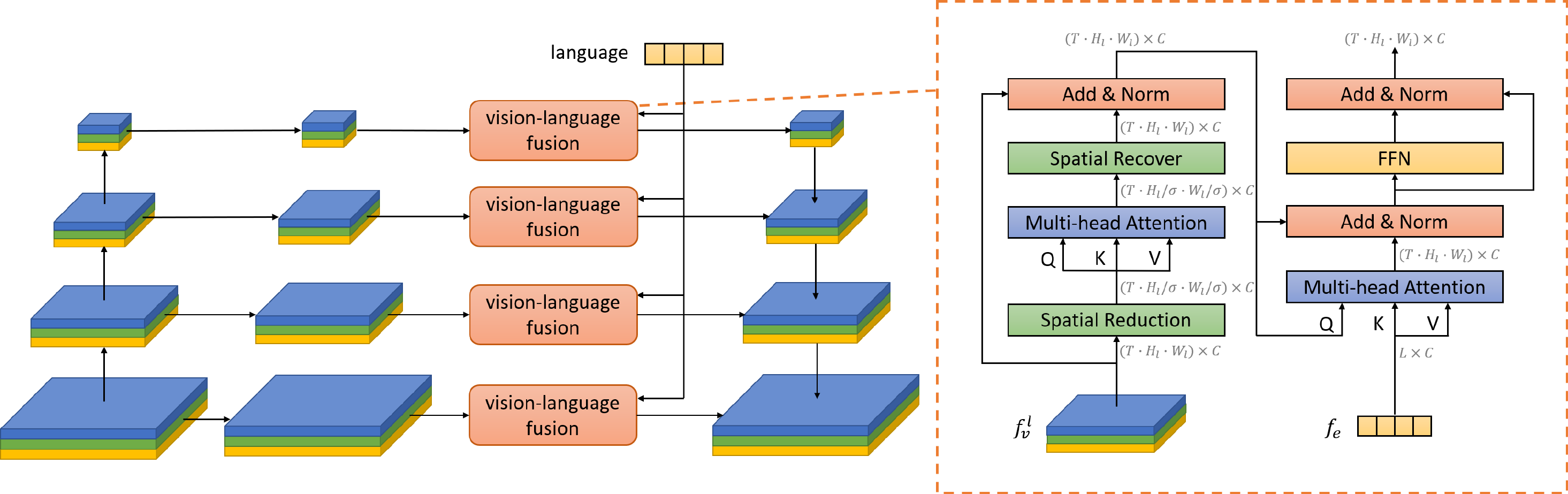}
\end{center}
\vspace{-4mm}
   \caption{The architecture of cross-modal feature pyramid network (CM-FPN). Note that different colors in the feature maps represent different frames. The visual and textual features are interacted in all the levels of feature maps. The vision-language fusion process is illustrated in the dash box on the right.}
\label{fig:fpn}
\end{figure*}

\myparagraph{Dynamic Convolution.} Suppose now we have obtained the semantically-rich feature maps $\mathcal{F}_{\text{seg}} = \left \{ f_{seg}^{t} \right \}_{t=1}^{T}$ (will be discussed in Sec. \ref{sec:cm-fpn}) for each frame, the question is how we perform the instance sequence segmentation and obtain the masks of referred object from them. Since the dynamic kernels have captured the object-level information, we use them as convolution filters on the feature maps for instance decoding. Considering that the location prior of dynamic kernels $\Omega$ provides a strong and robust reference for the referred object, we concatenate the feature maps $\mathcal{F}_{seg}$ with relative coordinates for each dynamic kernel. Finally, the binary segmentation masks are generated by performing dynamic convolution between the conditional convolutional weights and their corresponding feature maps:

\begin{equation}
    \label{eq:dynamic_convolution}
    \hat{s}_{i} = \left \{ \hat{f}_{i} \circledast \omega_{i} \right \}_{i=1}^{N_{q}}
\end{equation}

\noindent where $\omega_{i}$ and $\hat{f}_{i}$ are the $i$-th dynamic kernel weights and its exclusive feature map, respectively. We reshape the output masks in frame-order sequence, resulting in a set as $\hat{\mathcal{S}} \in \mathbb{R}^{T \times N \times \frac{H}{4} \times \frac{W}{4}}$.

\myparagraph{Illustration of conditional queries.} It is well known that the decoder embedding and position embedding in Transformer decoder encode the content and spatial information respectively. In our framework, these two parts are fed with the text sentence feature and learanble queries parameters, so that all the queries are restricted by the language expression. As shown in figure \ref{fig:vis_box}, these queries will focus on the referred object only even if other objects exist in the video. And there will be one query with much higher score while the scores of other queries will be suppressed.

\subsection{Cross-modal Feature Pyramid Network} \label{sec:cm-fpn}


Feature pyramid network (FPN) ~\cite{lin2017fpn} is adopted to produce multi-scale feature maps for video frames. We construct a 4-level pyramid with the spatial stride from 4$\times$ to 32$\times$. Specifically, the first three stage features of Transformer encoded memory (with spatial strides $\left \{ 8, 16, 32 \right \}$) and the 4$\times$ feature from visual backbone are stacked to form the hierarchical features. Although the standard FPN can already provide a high-resolution feature map with rich visual semantics, such feature map lacks the linguistic information and would be sub-optimal for the cross-modal task. The previous work ~\cite{seo2020urvos} only incorporates the language feature on the top level of FPN, which is a coarse fusion fashion. Here, we design a cross-modal feature pyramid network (CM-FPN) to perform multi-scale cross-modal fusion for finer interaction, as shown in Figure \ref{fig:fpn}.

In each level, the interaction process is achieved by the vision-language fusion module. And we take the $l$-th level feature of FPN as an example to clarify the process. Here, we use $f_{v}^{l} \in \mathbb{R}^{ T \times H_{l} \times W_{l} \times C}$ to represent the $l$-th level visual feature for simplicity. To model the spatio-temporal pixel-level relations of vision feature, we expect to feed it into a multi-head self-attention (MHSA) module. However, the computation of dense similarities make it intractable for the high-resolution feature maps. Inspired by ~\cite{wang2021pvt, fan2021mvit}, we propose the \textit{spatial reduction} and \textit{spatial recover} operations to address the issue. Before the MHSA module, the spatial size of vision feature $F_{v}^{l}$ is downsampled by a factor of $\sigma$ while the temporal dimension is kept unchanged. Thus, the complexity of self-attention ~\cite{vaswani2017transformer} operation would be greatly reduced, making the fusion module can be inserted into each level of FPN. Then, the spatial size of vision feature is recovered to $H_{l} \times W_{l}$ for maintaining fine-grained information. We set the downsample factors as $[8, 4, 2, 1]$ for the 4-level features maps. 
Next, $f_{v}^{l}$ interact with word-level feature $f_{e}$ in a \textit{cross-attention} way, where the \textit{query}, \textit{key} are vision and language feature, respectively:

\begin{equation}
    \label{eq:interact}
    \text{Interact}(f_{v}^{l}, f_{e})= \text{Softmax}(\frac{f_v^{l}W^{Q} \cdot (f_{e}W^{K})^{\text{T}}}{\sqrt{d_{\text{head}}}})f_{e}W^{V}
\end{equation}

\noindent where $W^{Q},W^{K},W^{V} \in \mathbb{R}^{C \times d_{\text{head}}}$ are learnable parameters. The visual feature plays the role of \textit{query} in attention mechanism, and thus the pixels on the feature map that are strongly related to the language expression will be strengthened. We upsample and sum the cross-modal feature maps following the standard FPN top-down structure. Finally, we apply an additional $3 \times 3$ convolutional layer on the feature maps with spatial stride 4 to get the final feature maps $\mathcal{F}_{seg} = \left \{ f_{\text{seg}}^{t} \right \}_{t=1}^{T}$, where $f_{\text{seg}}^{t} \in \mathbb{R}^{\frac{H}{4} \times \frac{W}{4} \times C_{d}}$.

\subsection{Instance Sequence Matching and Loss} \label{sec:instance_matching}

Using $N$ \textit{conditional queries}, we generate the set of $N_{q} = T \times N$ predictions, which can be regarded as the trajectories of $N$ instances on $T$ frames. As described previous, the predictions across frames maintain the same relative positions. Therefore, we can supervise the instance sequence as a whole using instance matching strategy ~\cite{wang2021vistr}. Let us denote the prediction set as $\hat{y} = \left \{  \hat{y}_{i}  \right \}_{i=1}^{N}$, and the predictions for the $i$-th instance is represented by:

\begin{equation}
    \label{eq:hatyi}
    \vspace{-2mm}
    \hat{y}_{i} = \left \{ \hat{p}_{i}^{t}, \hat{b}_{i}^{t}, \hat{s}_{i}^{t} \right \}_{t=1}^{T}
\end{equation}

\noindent For the $t$-th frame, $\hat{p}_{i}^{t} \in \mathbb{R}^{1}$ is a probability scalar indicating whether the instance corresponds to the referred object and this object is visible in the current frame. $\hat{b}_{i}^{t} \in \mathbb{R}^{4}$ is the normalized vector defines the center coordinates as well as the height and width of predicted box. $\hat{s}_{i}^{t} \in \mathbb{R}^{\frac{H}{4} \times \frac{W}{4}}$ is the predicted binary segmentation mask.

Since there is only one referred object in the video, the ground-truth instance sequence is represented as $y = \left \{ c^{t}, b^{t}, s^{t} \right \}_{t=1}^{T}$. $c^{t}$ is an one-hot value and it equals 1 when the ground-truth instance is visible in the frame $I_{t}$ otherwise 0. To train the network, we first find the best prediction as the positive sample via minimizing the matching cost:

\begin{equation}
    \label{eq:min}
    \vspace{-2mm}
    \hat{y}_{\text{pos}}= \mathop{\arg\min}_{\hat{y}_{i}\in \hat{y}} \mathcal{L}_{match}(y,\hat{y}_{i})
    \vspace{-2mm}
\end{equation}

\noindent where

\begin{equation}
\label{eq:match}
\begin{aligned}
\vspace{-2mm}
    \mathcal{L}_{match}(y,\hat{y}_{i})  &=     
    \lambda_{cls}  \mathcal{L}_{cls}(y,\hat{y}_{i}) 
    + \lambda_{\text{box}} \mathcal{L}_{\text{box}}(y,\hat{y}_{i}) \\
    &+ \lambda_{\text{mask}}  \mathcal{L}_{\text{mask}}(y,\hat{y}_{i}) 
    \vspace{-2mm}
\end{aligned}
\end{equation}

The matching cost is computed from each frame and normalized by the frame number. Here, $\mathcal{L}_{cls}(y,\hat{y}_{i})$ is the focal loss ~\cite{lin2017retinanet} that supervises the predicted instance sequence reference results. The box-related loss sums up the L1 loss and GIoU loss ~\cite{rezatofighi2019giou}. And the mask-related loss is the combination of DICE loss ~\cite{milletari2016dice} and binary mask focal loss. Both the two mask losses are spatio-temporally calculated over the entire video clip. The network is optimized by minimizing the total loss $\mathcal{L}_{match}$ for positive samples while letting the negative samples predict the $\varnothing$ class.

\subsection{Inference}

As mentioned previously, ReferFormer can handle the videos of arbitrary length in a single forward pass since all the frames share the same initial \textit{conditional queries}. Given the video and language expression, ReferFormer will predict $N$ instance sequence. For each instance query, we average the predicted reference probabilities over all the frames and obtain the reference score set $\mathcal{P} = \left \{ p_i \right \}_{i=1}^{N}$. We select the instance sequence with the highest average score and its index is denoted as $\sigma$:

\begin{equation}
    \label{eq:inference}
    \vspace{-2mm}
    \sigma = \mathop{\arg\max}_{i \in \left \{ 1, 2, ...,N \right \}}p_{i}
    \vspace{2mm}
\end{equation}

The final segmentation masks for each frame $\mathcal{S}= \left \{ s_{t} \right \}_{t=1}^{T}$ is obtained from the mask candidates set $\hat{\mathcal{S}}$ by selecting the corresponding queries indexed with $\sigma$. No post-process is needed for associating objects since the linked queries naturally track the same instance.

\section{Experiments}

\subsection{Datasets and Metrics}

\myparagraph{Datasets.} The experiments are conducted on the four popular R-VOS benchmarks: Ref-Youtube-VOS ~\cite{seo2020urvos}, Ref-DAVIS17 ~\cite{khoreva2018rvos}, A2D-Sentences and JHMDB-Sentences ~\cite{gavrilyuk2018a2dsentences}. 
Ref-Youtube-VOS ~\cite{seo2020urvos} is a large-scale benchmark which covers 3,978 videos with $\sim$15K language descriptions. Ref-DAVIS17 ~\cite{khoreva2018rvos} is built upon DAVIS17 ~\cite{pont2017davis} by providing the language description for a specific object in each video and contains 90 videos. A2D-Sentences and JHMDB-Sentences are created by providing the additional textual annotations on the original A2D ~\cite{xu2015a2d} and JHMDB ~\cite{jhuang2013jhmdb} datasets. A2D-Sentences contains 3,782 videos and each video has 3-5 frames annotated with the pixel-level segmentation masks. JHMDB-Sentences has 928 videos with the 928 corresponding sentences in total. 

\myparagraph{Evaluation Metrics.} We use the standard evaluation metrics for Ref-Youtube-VOS and Ref-DAVIS17: region similarity ($\mathcal{J}$), contour accuracy ($\mathcal{F}$) and their average value ($\mathcal{J} \& \mathcal{F}$). For Ref-Youtube-VOS, as the annotations of validation set are not released publicly, we evaluate our method on the official challenge server
\footnote{\textcolor{magenta}{https://competitions.codalab.org/competitions/29139}}. Ref-DAVIS17 is evaluated by the official evaluation code \footnote{\textcolor{magenta}{https://github.com/davisvideochallenge/davis2017-evaluation}}.

On A2D-Sentences and JHMDB-Sentences, the model is evaluated with the criteria of Precision@K, Ovrall IoU, Mean IoU and mAP over 0.50:0.05:0.95. The Precision@K measures the percentage of test samples whole IoU scores are higher than the threshold K. Following standard protocol, the thresholds are set as 0.5:0.1:0.9.

\begin{table*}[t]
    \begin{center}
        \begin{tabular}{l | c | c c c | c c c}

\toprule

\multirow{2}{*}{Method} & \multirow{2}{*}{Backbone} & \multicolumn{3}{c}{Ref-Youtube-VOS} & \multicolumn{3}{c}{Ref-DAVIS17} \\

\arrayrulecolor{white}\cline{3-8}
\arrayrulecolor{black}\cline{3-8}
\arrayrulecolor{black}\cline{3-8}
\arrayrulecolor{black}\cline{3-8}
\arrayrulecolor{white}\cline{3-8}

 & & $\mathcal{J}\&\mathcal{F}$ & $\mathcal{J}$ & $\mathcal{F}$ & 
     $\mathcal{J}\&\mathcal{F}$ & $\mathcal{J}$ & $\mathcal{F}$ \\

\arrayrulecolor{white}\hline
\arrayrulecolor{black}\hline
\arrayrulecolor{white}\hline

\multicolumn{8}{l}{\textbf{Spatial Visual Backbones}} \\

\arrayrulecolor{white}\hline
\arrayrulecolor{black}\hline
\arrayrulecolor{white}\hline

CMSA ~\cite{ye2019cmsa} & ResNet-50 & 34.9 & 33.3 & 36.5 & 34.7 & 32.2 & 37.2 \\
CMSA + RNN ~\cite{ye2019cmsa}  & ResNet-50 & 36.4 & 34.8 & 38.1 & 40.2 & 36.9 & 43.5 \\ 
URVOS ~\cite{seo2020urvos} & ResNet-50 & 47.2 & 45.3 & 49.2 & 51.5 & 47.3 & 56.0 \\
ReferFormer & ResNet-50 & 55.6 & 54.8 & 56.5 & \textbf{58.5} & \textbf{55.8} & \textbf{61.3} \\
ReferFormer$^{*}$ & ResNet-50 & \textbf{58.7} & \textbf{57.4} & \textbf{60.1} & - & - & - \\

\arrayrulecolor{white}\hline
\arrayrulecolor{black}\hline
\arrayrulecolor{white}\hline

PMINet ~\cite{ding2021pminet} & ResNeSt-101 & 48.2 & 46.7 & 49.6 & - & - & - \\
PMINet + CFBI ~\cite{ding2021pminet} & ResNeSt-101 & 53.0 & 51.5 & 54.5 & - & - & - \\
CITD$^{*}$ ~\cite{liang2021topdown} & ResNet-101 & 56.4 & 54.8 & 58.1 & - & - & - \\
ReferFormer & ResNet-101 & 57.3 & 56.1 & 58.4 & - & - & - \\
ReferFormer$^{*}$ & ResNet-101 & \textbf{59.3} & \textbf{58.1} & \textbf{60.4} & - & - & -  \\          

\arrayrulecolor{white}\hline
\arrayrulecolor{black}\hline
\arrayrulecolor{white}\hline

PMINet + CFBI ~\cite{ding2021pminet} & Ensemble & 54.2 & 53.0 & 55.5 & - & - & - \\
CITD ~\cite{liang2021topdown} & Ensemble & 61.4 & 60.0 & 62.7 & - & - & - \\
ReferFormer & Swin-L & 62.4 & 60.8 & 64.0 & \textbf{60.5} & \textbf{57.6} & \textbf{63.4} \\
ReferFormer$^{*}$ & Swin-L & \textbf{64.2} & \textbf{62.3} & \textbf{66.2} & - & - & - \\

\arrayrulecolor{white}\hline
\arrayrulecolor{black}\hline
\arrayrulecolor{white}\hline

\multicolumn{8}{l}{\textbf{Spatio-temporal Visual Backbones}} \\

\arrayrulecolor{white}\hline
\arrayrulecolor{black}\hline
\arrayrulecolor{white}\hline

MTTR$^{\dag}$ ($\omega=12$)  ~\cite{botach2021mttr} & \multirow{2}{*}{Video-Swin-T} & 55.3 & 54.0 & 56.6 & - & - & - \\
ReferFormer $^{\dag}$ ($\omega=5$) &  & \textbf{56.0} & \textbf{54.8} & \textbf{57.3} & - & - & - \\

\arrayrulecolor{white}\hline
\arrayrulecolor{black}\hline
\arrayrulecolor{white}\hline

ReferFormer & \multirow{2}{*}{Video-Swin-T} & 59.4 & 58.0 & 60.9 & - & - & - \\
ReferFormer$^{*}$ & & \textbf{62.6} & \textbf{59.9} & \textbf{63.3} & - & - & - \\

\arrayrulecolor{white}\hline
\arrayrulecolor{black}\hline
\arrayrulecolor{white}\hline

ReferFormer & \multirow{2}{*}{Video-Swin-S} & 60.1 & 58.6 & 61.6 & - & - & - \\
ReferFormer$^{*}$ & & \textbf{63.3} & \textbf{61.4} & \textbf{65.2} & - & - & - \\

\arrayrulecolor{white}\hline
\arrayrulecolor{black}\hline
\arrayrulecolor{white}\hline

ReferFormer & \multirow{2}{*}{Video-Swin-B} & 62.9 & 61.3 & 64.6 & \textbf{61.1} & \textbf{58.1} & \textbf{64.1} \\
ReferFormer$^{*}$ & & \textbf{64.9} & \textbf{62.8} & \textbf{67.0} & - & - & - \\

\arrayrulecolor{white}\hline
\arrayrulecolor{black}\hline
\arrayrulecolor{white}\hline

\end{tabular}
    \end{center}
    \vspace{-4mm}
    \caption{Comparison with the state-of-the-art methods on Ref-Youtube-VOS and Ref-DAVIS17. $^{*}$ means joint trainig with Ref-COCO dataset. $\dag$ indicates the spatio-temporal visual backbone is trained from scratch.}
    \label{tab:main_table}
    \vspace{-2mm}
\end{table*}

\subsection{Implementation Details.}

\myparagraph{Model Settings.} We test our models under different visual backbones including: ResNet ~\cite{he2016resnet}, Swin Transformer ~\cite{liu2021swin} and Video Swin Transformer ~\cite{liu2021videoswin}. The text encoder is selected as RoBERTa ~\cite{liu2019roberta} and its parameters are frozen during the entire training stage. Following ~\cite{zhu2020deformable}, we use the last stage features from the visual backbone as the input to Transformer, their corresponding spatial strides are $\left \{ 8, 16, 32 \right \}$. In the Transformer model, we adopt 4 encoder layers and 4 decoder layers and the hidden dimension is $C=256$. The number of \textit{conditional query} is set as 5 otherwise specified.

\myparagraph{Training Details.} During training, we use sliding-windows to obtain the clips from a video and each clip consist of 5 randomly sampled frames. Following ~\cite{wang2021vistr}, the data augmentation includes random horizontal flip, random resize, random crop and photometric distortion. All frames are downsampled so that the short side has the size of 360 and the maximum size for the long side is 640 to fit GPU memory. The coefficients for losses are set as $\lambda_{cls}=2$, $\lambda_{L1}=5$, $\lambda_{giou}=2$, $\lambda_{dice}=5$, $\lambda_{focal}=2$.

Most of our experiments follow the pretrain-then-finetune process. And some models are trained from scratch for fair comparison. Additionally, on Ref-Youtube-VOS, we also reports the results by training the mixed data from Ref-Youtube-VOS and Ref-COCO ~\cite{yu2016refcoco}. The joint training technique has proven the effectiveness in many VIS tasks ~\cite{athar2020stem-seg, lin2021propose-reduce, wu2021seqformer}. Please see more in the supplementary materials.



\myparagraph{Inference Details.} During inference, the video frames are downscaled to 360p. We directly output the predicted segmentation masks without post-process. On Ref-Youtube-VOS, we further use a simple post-process technique to refine the object masks. Concretely, we first select a frame with the highest prediction score as the reference frame. Then, we apply the off-the-shelf mask propagation method CFBI ~\cite{yang2020cfbi} to propagate the predicted mask of this frame forward and backward to the entire video. The results with post-process are shown in Table \ref{tab:ab_backbone}.

\subsection{Main Results}

\begin{table*}[t]
    \begin{center}
        \begin{tabular}{l | c | c c c c c | c c | c}

\toprule

\multirow{2}{*}{Method} & \multirow{2}{*}{Backbone} & \multicolumn{5}{c |}{Precision} & \multicolumn{2}{c |}{IoU} & \multirow{2}{*}{mAP} \\

 & & P@0.5 & P@0.6 & P@0.7 & P@0.8 & P@0.9 & Overall & Mean &  \\

\arrayrulecolor{white}\hline
\arrayrulecolor{black}\hline
\arrayrulecolor{white}\hline

Hu \etal ~\cite{hu2016segmentation} & VGG-16 & 34.8 & 23.6 & 13.3 & 3.3 & 0.1 & 47.4 & 35.0 & 13.2 \\
Gavrilyuk \etal ~\cite{gavrilyuk2018a2dsentences}  & I3D & 47.5 & 34.7 & 21.1 & 8.0 & 0.2 & 53.6 & 42.1 & 19.8 \\ 
CMSA + CFSA ~\cite{ye2021cfsa} & ResNet-101 & 48.7 & 43.1 & 35.8 & 23.1 & 5.2 & 61.8 & 43.2 & - \\
ACAN ~\cite{wang2019acan} & I3D & 55.7 & 45.9 & 31.9 & 16.0 & 2.0 & 60.1 & 49.0 & 27.4 \\
CMPC-V ~\cite{liu2021cmpc} & I3D & 65.5 & 59.2 & 50.6 & 34.2 & 9.8 & 65.3 & 57.3 & 40.4 \\
ClawCraneNet ~\cite{liang2021clawcranenet} & ResNet-50/101 & 70.4 & 67.7 & 61.7 & 48.9 & 17.1 & 63.1 & 59.9 & - \\
MTTR ($\omega = 8$) ~\cite{botach2021mttr} & Video-Swin-T & 72.1 & 68.4 & 60.7 & 45.6 & 16.4 & 70.2 & 61.8 & 44.7 \\
MTTR ($\omega = 10$) ~\cite{botach2021mttr} & Video-Swin-T & 75.4 & 71.2 & 63.8 & 48.5 & 16.9 & 72.0 & 64.0 & 46.1 \\
ReferFormer$^{\dag}$ ($\omega = 6$) & Video-Swin-T & 76.0 & 72.2 & 65.4 & 49.8 & 17.9 & 72.3 & 64.1 & 48.6 \\

\arrayrulecolor{white}\hline
\arrayrulecolor{black}\hline
\arrayrulecolor{white}\hline

ReferFormer ($\omega = 5$) & Video-Swin-T & 82.8 & 79.2 & 72.3 & 55.3 & 19.3 & 77.6 & 69.6 & 52.8 \\

ReferFormer ($\omega = 5$) & Video-Swin-S & 82.6 & 79.4 & 73.1 & 57.4 & 21.1 & 77.7 & 69.8 & 53.9 \\

\textbf{ReferFormer} ($\omega = 5$) & Video-Swin-B & \textbf{83.1} & \textbf{80.4} & \textbf{74.1} & \textbf{57.9} & \textbf{21.2} & \textbf{78.6} & \textbf{70.3} & \textbf{55.0} \\

\arrayrulecolor{white}\hline
\arrayrulecolor{black}\hline
\arrayrulecolor{white}\hline

\end{tabular}
    \end{center}
    \vspace{-4mm}
    \caption{Comparison with the state-of-the-art methods on A2D-Sentences. $^{\dag}$ means our model is trained from scratch.}
    \label{tab:a2d_table}
    \vspace{-2mm}
\end{table*}

\begin{table*}[t]
    \begin{center}
        \begin{tabular}{l | c | c c c c c | c c | c}

\toprule

\multirow{2}{*}{Method} & \multirow{2}{*}{Backbone} & \multicolumn{5}{c |}{Precision} & \multicolumn{2}{c |}{IoU} & \multirow{2}{*}{mAP} \\

 & & P@0.5 & P@0.6 & P@0.7 & P@0.8 & P@0.9 & Overall & Mean &  \\

\arrayrulecolor{white}\hline
\arrayrulecolor{black}\hline
\arrayrulecolor{white}\hline

Hu \etal ~\cite{hu2016segmentation} & VGG-16 & 63.3 & 35.0 & 8.5 & 0.2 & 0.0 & 54.6 & 52.8 & 17.8 \\
Gavrilyuk \etal ~\cite{gavrilyuk2018a2dsentences}  & I3D & 69.9 & 46.0 & 17.3 & 1.4 & 0.0 & 54.1 & 54.2 & 23.3 \\ 
CMSA + CFSA ~\cite{ye2021cfsa} & ResNet-101 & 76.4 & 62.5 & 38.9 & 9.0 & 0.1 & 62.8 & 58.1 & - \\
ACAN ~\cite{wang2019acan} & I3D & 75.6 & 56.4 & 28.7 & 3.4 & 0.0 & 57.6 & 58.4 & 28.9 \\
CMPC-V ~\cite{liu2021cmpc} & I3D & 81.3 & 65.7 & 37.1 & 7.0 & 0.0 & 61.6 & 61.7 & 34.2 \\
ClawCraneNet ~\cite{liang2021clawcranenet} & ResNet-50/101 & 88.0 & 79.6 & 56.6 & 14.7 & 0.2 & 64.4 & 65.6 & - \\
MTTR ($\omega = 8$) ~\cite{botach2021mttr} & Video-Swin-T & 91.0 & 81.5 & 57.0 & 14.4 & 0.1 & 67.4 & 67.9 & 36.6 \\
MTTR ($\omega = 10$) ~\cite{botach2021mttr} & Video-Swin-T & 93.9 & 85.2 & 61.6 & 16.6 & 0.1 & 70.1 & 69.8 & 39.2 \\
ReferFormer$^{\dag}$ ($\omega = 6$) & Video-Swin-T & 93.3 & 84.2 & 61.4 & 16.4 & 0.3 & 70.0 & 69.3 & 39.1 \\

\arrayrulecolor{white}\hline
\arrayrulecolor{black}\hline
\arrayrulecolor{white}\hline

ReferFormer ($\omega = 5$) & Video-Swin-T & 95.8 & 89.3 & 66.8 & 18.9 & 0.2 & 71.9 & 71.0 & 42.2 \\

ReferFormer ($\omega = 5$) & Video-Swin-S & 95.8 & 90.1 & 68.7 & 20.3 & 0.2 & 72.8 & 71.5 & 42.4 \\

\textbf{ReferFormer} ($\omega = 5$) & Video-Swin-B & \textbf{96.2} & \textbf{90.2} & \textbf{70.2} & \textbf{21.0} & \textbf{0.3} & \textbf{73.0} & \textbf{71.8} & \textbf{43.7} \\

\arrayrulecolor{white}\hline
\arrayrulecolor{black}\hline
\arrayrulecolor{white}\hline

\end{tabular}
    \end{center}
    \vspace{-4mm}
    \caption{Comparison with the state-of-the-art methods on JHMDB-Sentences. $^{\dag}$ means our model is trained from scratch.}
    \label{tab:jhmdb_table}
    \vspace{-2mm}
\end{table*}

\myparagraph{Ref-Youtube-VOS \& Ref-DAVIS17 } 
We compare our method with other state-of-the-art methods in Table \ref{tab:main_table}. CITD ~\cite{liang2021topdown} and PMINet ~\cite{ding2021pminet} are the top-2 solutions in \texttt{2021 Ref-Youtube-VOS Challenge}. Their ensemble results are based on building 5 and 4 models, respectively. It can be observed that ReferFormer outperforms previous methods on the two datasets under all metrics and with a large marge. On Ref-Youtube-VOS, ReferFormer with a ResNet-50 backbone achieves the overall $\mathcal{J}\&\mathcal{F}$ of 55.6, which is 8.4 points higher than the previous state-of-the-art work URVOS ~\cite{seo2020urvos}, and even beats PMINet ~\cite{ding2021pminet} using the ensemble models and adopting post-process (55.6 vs 54.2). Using the strong Swin-Large ~\cite{liu2021swin} backbone, ReferFormer reaches the surprising 62.4 $\mathcal{J}\&\mathcal{F}$ without bells and whistles, which obviously exceeds the ensemble results of the complicated, multi-stage method CITD ~\cite{liang2021topdown}. By using the joint training process, the performance of our model can be further boosted to 64.2 $\mathcal{J}\&\mathcal{F}$, creating a fairly high new record. Additionally, we also test the Video Swin Transformer ~\cite{liu2021videoswin} as the backbones. It is well known that the spatio-temporal visual encoder has strong ability to capture both the spatial characteristics and the temporal cues. For a fair comparison with MTTR ~\cite{botach2021mttr}, we train our model with the Video-Swin-Tiny backbone from scratch. It can be seen that our method outperforms MTTR under all the metrics with the smaller window size (5 vs 12). Comparing the results of ReferFormer under Video-Swin-Tiny backbone, it proves that the model benefits from the pretraining stage and joint training process to address the overfitting issue.

On Ref-DAVIS17, our method also achieves the best results under the same ResNet-50 setting (58.5 $\mathcal{J}\&\mathcal{F}$). And the performance consistently improves by using stronger backbones, 
which proves the generality of our method.

\vspace{-2mm}
\myparagraph{A2D-Sentences \& JHMDB-Sentences } 
We further evaluate our method on the A2D-Sentences dataset and compare the performance with other state-of-the-art methods in Table \ref{tab:a2d_table}. ClawCraneNet ~\cite{liang2021clawcranenet} is a mutli-stage method which use the off-the-shelf instance segmentation model (with ResNet-101 backbone) to provide the mask candidates. From Table \ref{tab:a2d_table}, it is obvious that our method achieves the impressive improvement over the previous methods. Compared with the recent MTTR ~\cite{botach2021mttr}, our method exhibits the clear performance advantange ($+$2.5 mAP) with smaller window size (6 vs. 10). Incorporating the pretraining stage, ReferFormer with Video-Swin-Base visual backbone achieves 55.0 mAP which shows a significant gain of 8.9 mAP over previous best result. And ReferFormer also demonstrates its strong ability to produce high-quality masks via the stringent metrics (\eg, 57.9 for P@0.8 and 21.2 for P@0.9). 

We also evaluate the models on JHMDB-Sentences without finetuning to further prove the generality of our method. As shown in Table \ref{tab:jhmdb_table}, ReferFormer significantly outperforms all the existing methods. It is noticeable that all the methods produce low scores on P@0.9. A possible reason is that the ground-truth masks are generated from human puppets, leading to the inaccurate mask annotations.

\begin{table}[t]
    \begin{center}
        \begin{tabular}{l | l l }

\toprule

Components &  \multicolumn{1}{c}{$\mathcal{J}$} & \multicolumn{1}{c}{$\mathcal{F}$} \\

\arrayrulecolor{white}\hline
\arrayrulecolor{black}\hline
\arrayrulecolor{white}\hline

\arrayrulecolor{white}\hline
\arrayrulecolor{black}\hline
\arrayrulecolor{white}\hline

Baseline & 47.2\negacc{$-$7.6} & 50.1\negacc{$-$7.2} \\

\arrayrulecolor{white}\hline
\arrayrulecolor{black}\hline
\arrayrulecolor{white}\hline

w/o Visual-language Fusion & 53.0\negacc{$-$1.8} & 56.2\negacc{$-$1.1} \\
w/o Relative Coordinates & 53.7\negacc{$-$1.1} & 55.9\negacc{$-$1.4} \\

\arrayrulecolor{white}\hline
\arrayrulecolor{black}\hline
\arrayrulecolor{white}\hline

Full Model & 54.8 & 57.3 \\

\arrayrulecolor{white}\hline
\arrayrulecolor{black}\hline
\arrayrulecolor{white}\hline

\end{tabular}
    \end{center}
    \vspace{-4mm}
    \caption{Ablation study on the components of ReferFormer. The visual backbone is Video-Swin-Tiny.}
    \label{tab:ab_component}
    \vspace{-5mm}
\end{table}

\subsection{Ablation Study}

In this section, we perform extensive ablation studies on Ref-Youtube-VOS to study the effect of core components in our model. All models are based on Video-Swin-Tiny visual backbone and we train the models from scratch otherwise specified. The detailed analysis is as follows.

\myparagraph{Component Analysis.} We build a simple Transformer bottom-up baseline. Specifically, considering a video clip of $T$ frames, we flatten the temporal and spatial dimension into one dimension and then concatenate the visual features with the textual features along length dimension to form the multi-modal feature map $f_{m} \in \mathbb{R}^{(T \times H \times W + L) \times C}$. The vanilla Transformer encoder builds the global dependencies between the visual and textual features. Afterwards, we extract the visual features from the encoded memory and construct a standard FPN-like architecture upon them for generating the segmentation masks. The baseline method operates the fixed-length video of 5 frames during the training and inference phases. We report the performance of the baseline method and also study the effect of core components in Table \ref{tab:ab_component}. 

First, from the first row of Table \ref{tab:ab_component}, the baseline method only achieves 47.2 $\mathcal{J}$ and 50.1 $\mathcal{F}$. This inferior behavior attributes to two reasons: (1) The baseline method can not distinguish the similar objects that are close together and tends to segment the most salient region. In contrast, our method performs well with only 1 conditional query (see Table \ref{tab:ablation}(a)), proving that dynamic convolution is essential for segmenting the referred object. (2) Our method uses a set of shared queries to track instances in all frames, and the best query is determined by the \textit{voting} scores of each frame. In this sense, our model can produce a reliable reasoning result and keep the temporal consistency in the entire video. On the contrary, the baseline method could be regarded as a image-level method that independently predicts the results of each frame even though the model is able to aggregate the information from other frames.

\begin{table}[t]
    \begin{center}
        \begin{tabular}{l | l l l }

\toprule

Backbone & \multicolumn{1}{c}{$\mathcal{J}\&\mathcal{F}$} & \multicolumn{1}{c}{$\mathcal{J}$} & \multicolumn{1}{c}{$\mathcal{F}$} \\

\arrayrulecolor{white}\hline
\arrayrulecolor{black}\hline
\arrayrulecolor{white}\hline

ResNet-50 & 55.6 & 54.8 & 56.5 \\
ResNet-50$^{*}$ & 59.4\posacc{$+$3.8} & 58.1\posacc{$+$3.3} & 60.8\posacc{$+$4.3} \\

\arrayrulecolor{white}\hline
\arrayrulecolor{black}\hline
\arrayrulecolor{white}\hline

ResNet-101 & 57.3 & 56.1 & 58.4 \\
ResNet-101$^{*}$ & 60.3\posacc{$+$3.0} & 58.8\posacc{$+$2.7} & 61.8\posacc{$+$3.4} \\

\arrayrulecolor{white}\hline
\arrayrulecolor{black}\hline
\arrayrulecolor{white}\hline

Swin-T & 58.7 & 57.6 & 59.9 \\
Swin-T$^{*}$ & 61.2\posacc{$+$2.5} & 59.7\posacc{$+$2.1} & 62.6\posacc{$+$2.7} \\

\arrayrulecolor{white}\hline
\arrayrulecolor{black}\hline
\arrayrulecolor{white}\hline

Swin-S & 59.6 & 58.1 & 61.1 \\
Swin-S$^{*}$ & 61.3\posacc{$+$1.7} & 59.7\posacc{$+$1.6} & 63.0\posacc{$+$1.9} \\

\arrayrulecolor{white}\hline
\arrayrulecolor{black}\hline
\arrayrulecolor{white}\hline

Swin-B & 61.8 & 60.1 & 63.4 \\
Swin-B$^{*}$ & 63.1\posacc{$+$1.3} & 61.4\posacc{$+$1.3} & 64.8\posacc{$+$1.4} \\

\arrayrulecolor{white}\hline
\arrayrulecolor{black}\hline
\arrayrulecolor{white}\hline

Swin-L & 62.4 & 60.8 & 64.0 \\
Swin-L$^{*}$ & 63.3\posacc{$+$0.9} & 61.6\posacc{$+$0.8} & 65.1\posacc{$+$1.1} \\
 
\arrayrulecolor{white}\hline
\arrayrulecolor{black}\hline
\arrayrulecolor{white}\hline

\end{tabular}
    \end{center}
    \vspace{-4mm}
    \caption{Ablation study on the visual backbones. * indicates using CFBI ~\cite{yang2020cfbi} as post-process.}
    \label{tab:ab_backbone}
    \vspace{-3mm}
\end{table}



\begin{figure*}[t]
\begin{center}
   \includegraphics[width=1.0\linewidth]{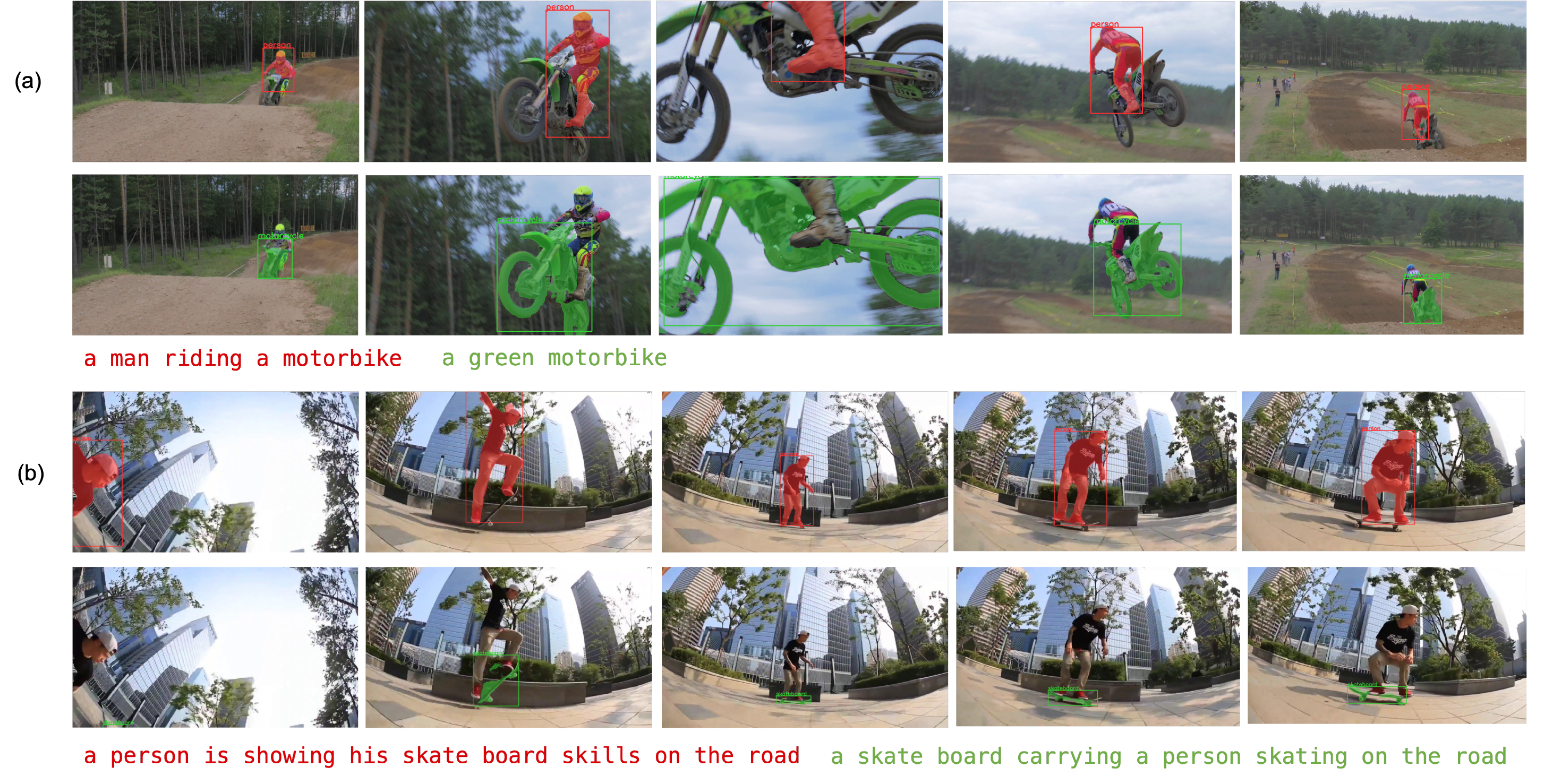}
\end{center}
\vspace{-5mm}
\caption{Visualization results on (a) Ref-DAVIS17 and (b) Ref-Youtube-VOS. Our unified framework is able to detect, segment and track the referred object simultaneously.}
\label{fig:visualize}
\end{figure*}

\begin{table*}[t]
\centering
    \begin{minipage}{0.3\textwidth} 
    \centering 
    {
        \setlength{\tabcolsep}{1.2mm}
\begin{tabular}
{c c c c}
\toprule
Queries & $\mathcal{J}\&\mathcal{F}$ & $\mathcal{J}$ & $\mathcal{F}$  \\
\midrule
1 & 53.6 & 52.7 & 54.5 \\
3 & 54.2 & 53.2 & 55.2 \\
5 & 56.0 & 54.8 & 57.3 \\
8 & 55.3 & 54.1 & 56.6 \\
\bottomrule
\end{tabular}
    }
    \end{minipage}
    \quad
    \begin{minipage}{0.3\textwidth} 
    \centering 
    {
        \setlength{\tabcolsep}{1.2mm}
\begin{tabular}
{c c c c}
\toprule
Frames & $\mathcal{J}\&\mathcal{F}$ & $\mathcal{J}$ & $\mathcal{F}$  \\
\midrule
1 & 50.0 & 48.4 & 51.6 \\
3 & 54.8 & 53.6 & 56.0 \\
5 & 56.0 & 54.8 & 57.3 \\
\bottomrule

\end{tabular}
    }
    \end{minipage}
    \quad
    \begin{minipage}{0.3\textwidth} 
    \centering 
    {
        \setlength{\tabcolsep}{1.2mm}
\begin{tabular}
{c c c c c c}
\toprule
Class & Box & Mask & $\mathcal{J}\&\mathcal{F}$ & $\mathcal{J}$ & $\mathcal{F}$  \\
\midrule

\checkmark & \checkmark & & 55.2 & 54.0 & 56.4 \\
\checkmark & & \checkmark & 54.5 & 53.5 & 55.5 \\
\checkmark & \checkmark & \checkmark & 56.0 & 54.8 & 57.3 \\

\bottomrule

\end{tabular}
    }
    \end{minipage}
\vspace{2mm} 
\\
\quad
    \begin{minipage}{0.3\textwidth} \raggedright
        (a) The effect of query number.
    \end{minipage}
    \quad
    \begin{minipage}{0.3\textwidth} \raggedright
        (b) The effect of frame number.
    \end{minipage}
    \quad
    \begin{minipage}{0.3\textwidth} \raggedright
        (c) The effect of label assignment method.
    \end{minipage}
\vspace{-1mm}
\\
\caption{Ablation study on different settings of ReferFormer. All the models are using Video-Swin-Tiny as visual backbone.}
\label{tab:ablation} 
\vspace{-0.2in}
\end{table*}

Second, comparing the second and last row of Table \ref{tab:ab_component}, we can see that the standard FPN has already achieved strong performance and the vision-language fusion process further helps to provide more accurate segmentation. This is because the object mask would be inaccurate due to light variation, whereas the cross-modal fusion uses the text as a complementary to strengthen the object pixel features and thus facilitates the segmentation prediction. Another technique is concatenating the relative coordinates of dynamic kernels with the mask features, this would help the model better determine the location of referred object and lead to performance improvement, as shown in the third row in Table \ref{tab:ab_component}.

\myparagraph{Visual Backbone.} We implement different visual backbones and report the results in Table \ref{tab:ab_backbone}. As expected, the performance of our model consistently increases by using stronger backbones. And the CFBI ~\cite{yang2020cfbi} post-process can help to further boost the performance under all backbone settings. Interestingly, we observe that the performance improvement by post-process tends to narrow when the backbone gets stronger, \eg, +3.8 for ResNet-50 and +0.9 for Swin-Large when considering the $\mathcal{J} \& \mathcal{F}$ metric. This phenomenon shows that the visual encoder is essential for providing reliable reasoning on which object is described and generating the precise masks.


\myparagraph{Number of Conditional Queries.} Benefit from the design of \textit{conditional queries}, all the initial object queries tend to find the referred object only. In this situation, we can only use a relatively small number of queries. In Table \ref{tab:ablation}(a), we study the effect of query number for each frame. It can be seen that the model achieves considerable results under all these settings, even with $N=1$. Certainly, more queries enable the model make judgement from a wide range of instance candidates, which could better handle the complicated scenes where the similar objects are clustered together. The performance saturates at $N=5$ and begins to slightly decrease when the query number gets larger. We conjecture that it is caused by the imbalance of label assignment as there is only one positive sample in each frame.

\myparagraph{Number of Training Clip Frames.} We study the effect of training clip frame number in Table \ref{tab:ablation}(b). Note that under $T=1$, the model can be viewed as an image-level method and the performance of metric $\mathcal{J}\&\mathcal{F}$ is only 50.0. When the frame number increases to 3, the model enjoys an significant $\mathcal{J}\&\mathcal{F}$ gain of 4.8. This is because using more frames to form a clip helps the model better aggregate the temporal action-related information. We choose $T=5$ by default.

\myparagraph{Label Assignment Method.} Our framework is able to predict the reference probability, box location and segmentation mask of the referred object. We find the optimal positive sample by minimizing the overall matching cost in Eq.\ref{eq:min}. There are some variants in the label assignment method and we carry out the comparison experiments in Table \ref{tab:ablation}(c). From the first two rows in Table \ref{tab:ablation}(c) we show that the lack of box or mask cost would both lead to the performance drop. With the segmentation-centric design, the mask cost is the most direct guidance for optimization, and the box provides the location prior for dynamic kernel. Thus, the combination of classification, box and mask cost shows more robustness.
\vspace{-2mm}

\subsection{Visualization Results}

We show the visualization results of our model in Figure \ref{fig:visualize}. It can be seen that ReferFormer is able to segment and track the referred object in challenging cases, \eg, person pose variations, instances occlusion and instances that are partially displayed or completely disappeared in the camera.
\vspace{-4mm}

\section{Conclusion}

In this work, we propose ReferFormer, an extremely simple and unified framework for referring video object segmentation. This framework provides a new perspective for the R-VOS task which views the language as queries. These queries are restricted to attend to the referred object only, and the object tracking is easily achieved by linking the corresponding queries. Given the video clip and an expression, our framework directly produces the segmentation masks as well as the detected boxes of the referred object in all frames without post-process. We validate our model on Ref-Youtube-VOS, Ref-DAVIS17, A2D-Sentences and JHMDB-Sentences and it shows the state-of-the-art performance on the four benchmarks.

{\small
\bibliographystyle{ieee_fullname}
\bibliography{main}
}

\clearpage

\appendix

\renewcommand\thefigure{\Alph{section}\arabic{figure}}  
\setcounter{figure}{0} 
\renewcommand\thetable{\Alph{section}\arabic{table}}  
\setcounter{table}{0}

\definecolor{codegreen}{rgb}{0,0.5,0}
\definecolor{codeblue}{rgb}{0.25,0.5,0.5}
\definecolor{codegray}{rgb}{0.6,0.6,0.6}

\lstset{
  backgroundcolor=\color{white},
  basicstyle=\fontsize{7.5pt}{8.5pt}\fontfamily{lmtt}\selectfont,
  columns=fullflexible,
  breaklines=true,
  captionpos=b,
  commentstyle=\fontsize{8pt}{9pt}\color{codegray},
  keywordstyle=\fontsize{8pt}{9pt}\color{codegreen},
  stringstyle=\fontsize{8pt}{9pt}\color{codeblue},
  frame=tb,
  otherkeywords = {self},
}

\section{Additional Dataset Details}

\textbf{Ref-Youtube-VOS} ~\cite{seo2020urvos} is a large-scale benchmark which covers 3,978 videos with $\sim$15K language descriptions. There are 3,471 videos with 12,913 expressions in training set and 507 videos with 2,096 expressions in validation set. According to the R-VOS competition, videos in the validation set are further split into 202 and 305 videos for the competition validation and test purpose. Since the test server is currently inaccessible, the results are reported by submitting our predictions to the validation server\footnote{\textcolor{magenta}{https://competitions.codalab.org/competitions/29139}}.

\textbf{Ref-DAVIS17} ~\cite{khoreva2018rvos} is built upon DAVIS17 ~\cite{pont2017davis} by providing the language description for a specific object in each video. It contains 90 videos with 1,544 expression sentences describing 205 objects in total. The dataset is split into 60 videos and 30 videos for training and validation, respectively. Since there are two annotators and each of them gives the \textit{first-frame} and \textit{full-video} textual description for one referred object, we report the results by averaging the scores using the official evaluation code \footnote{\textcolor{magenta}{https://github.com/davisvideochallenge/davis2017-evaluation}}. 
\section{Additional Implementation Details}

Our model is optimized using AdamW ~\cite{loshchilov2017adamw} optimizer with the weight decay of $5 \times 10^{-4}$, initial learning rate of $5 \times 10^{-5}$ for visual backbone and $10^{-4}$ for the rest. We first pretrain our model on the image referring segmentation datasets Ref-COCO ~\cite{yu2016refcoco}, Ref-COCOg ~\cite{yu2016refcoco} and Ref-COCO+ ~\cite{mao2016refcoco+} by setting $T=1$ with the batch size of 2 on each GPU. The pretrain procedure runs for 12 epochs with the learning rate decays divided by 10 at epoch 8 and 10. Then, on Ref-Youtube-VOS, we finetune the model for 6 epochs with 1 video clip per GPU. The learning rate decays by 10 at the 3-th and 5-th epoch. On Ref-DAVIS17, we directly report the results using the model trained on Ref-Youtube-VOS without finetune.

For A2D-Sentences, we feed the model with the window size of 5. The model is finetuned for 6 epochs with the learning rate decays at the 3-th and 5-th epoch by a factor of 0.1. On JHMDB-Sentences, following the previous works, we evaluate the generality of our method using the model trained on A2D-Sentences without finetune.

Additionally, on the Ref-Youtube-VOS, we also adopt the joint training technique by mixing the dataset with Ref-COCO/+/g. Specifically, for each image in the Ref-COCO dataset, we augment it with $\pm 20^\circ$ to form a 5-frame pseudo video clip. The joint training takes 12 epochs with the learning rate decays at the 8-th and 10-th epoch by a factor of 0.1. We use 32 V100 GPUS for the joint training and each GPU is fed with 2 video clips. It should be noted that the text encoder is froze all the time.

\section{Additional Details of Dynamic Convolution}

We give the pseudo-code of dynamic convolution in Figure \ref{fig:pesudo-code}, where we take one dynamic kernel for clarification. Specifically, a linear projection is applied to transform the instance embedding into dynamic convolutional weights. Then, the mask features pass through consecutive dynamic convolutional layers with the ReLU activation function. There is no normalization or activation after the last dynamic convolutional layer, and the output channel number of last layer is 1.

\begin{figure}[h]
\begin{lstlisting}[language=Python]
def dynamic_convolution(mask_feats, dynamic_feats):
    # mask_feats: (B, C, H/4, W/4)
    # dynamic_feats: (B, C)

    # parameters of dynamic convs: (B, num_params)
    dynamic_params = linear(dynamic_features)
    # parse conv parameters
    # weights[l]: (out_channels, in_channels, 1, 1)
    # bias[l]: (out_channels)
    weights, bias = parse_dynamic_params(dynaic_params)
    
    # dynamic convolution
    n_layer = len(weights)
    x = mask_feats
    for i, (x, b) in enumerate(zip(weights, bias)):
        x = conv2d(x, w, bias=b, stride=1, padding=0)
        if i < n_layer - 1:
            x = relu(x)
    # x: (B, H/4, W/4)
    return x
\end{lstlisting}
\vspace{-1em}
\caption{Pseudo-code of dynamic convolution, we take one dynamic kernel for clarification. For multiple dynamic kernels, we use group convolution in \texttt{conv2d} for efficient implementation. \texttt{linear}: linear projection. }
\label{fig:pesudo-code}
\vspace{-2mm}
\end{figure}

\section{Additional Experiment Results}

By default, our models are trained in the class-agnostic way, \ie, decide whether the object is referred or not. As described in Sec \ref{sec:language_as_queries}, the class head can be easily modified to predict the referred object category by simply change the class number. In this way, we train our model in a class-discriminative way and show the results in Table \ref{tab:ab_class}. We could observe the class-agnostic training method has clear performance gain ($+$2.1 $\mathcal{J}\&\mathcal{F}$) over the strong class-discriminative training results, since the binary classification is easier to optimize. The selection of training method can flexibly depend on the usage in real applications.

\begin{table}[H]
    \begin{center}
        \setlength{\tabcolsep}{1.5mm}
\begin{tabular}
{c c c c}
\toprule
Class Agnostic & $\mathcal{J}\&\mathcal{F}$ & $\mathcal{J}$ & $\mathcal{F}$  \\
\midrule
 & 53.9 & 52.8 & 55.0 \\
\checkmark & 56.0 & 54.8 & 57.3 \\
\bottomrule

\end{tabular}
    \end{center}
    \vspace{-4mm}
    \caption{Ablation study on the class-agnostic training.}
    \label{tab:ab_class}
    \vspace{-3mm}
\end{table}

\end{document}